\documentclass[a4paper,11pt,twocolumn,twoside]{article}
\usepackage{graphicx}
\usepackage{sepln_en}
\usepackage{fullname}
\usepackage[utf8]{inputenc}
\usepackage[english]{babel}
\usepackage{url}
\usepackage{tabularx}
\usepackage{booktabs}
\usepackage{multirow}
\usepackage[font=small,labelfont=bf]{caption}
\usepackage{makecell}
\usepackage{adjustbox}

\input epsf

\title{Evaluating Compact LLMs for Zero-Shot Iberian Language Tasks on End-User Devices}

\author {\textbf{Luis Couto Seller$^1$, Íñigo Sanz Torres$^1$, Adrián Vogel-Fernández$^1$,}\\\textbf{Carlos González Carballo$^1$, Pedro Miguel Sánchez Sánchez$^{1*}$,}\\\textbf{Adrián Carruana Martín$^1$, Enrique de Miguel Ambite$^1$}\\
$^1$Advantx Technological Foundation (Funditec), 28046-Madrid, Spain\\
$^*$Corresponding author: pmsanchezs@funditec.es\\
}

\seplntranstitle{Evaluación Comparativa de LLMs Compactos para Tareas de Lenguas Ibéricas en Dispositivos de Usuario}

\seplnclave{Grandes Modelos de Lenguaje (LLMs), Evaluación Comparativa, Lenguas Ibéricas, PNL Multilingüe, Modelos Compactos, Computación en el Borde}

\seplnresumen{Los Grandes Modelos de Lenguaje han avanzado significativamente el ámbito del procesamiento del lenguaje natural, logrando un rendimiento notable en generación de texto, la traducción y el razonamiento. Sin embargo, sus elevados requisitos computacionales restringen su despliegue a sistemas de alta gama, limitando su accesibilidad en dispositivos de consumo. Este desafío es especialmente pronunciado para idiomas con pocos recursos, como los hablados en la Península Ibérica, donde la disponibilidad relativamente limitada de recursos lingüísticos y benchmarks dificulta una evaluación efectiva. Este trabajo presenta una evaluación integral de modelos de lenguaje compactos y de última generación en varias tareas esenciales, adaptadas específicamente para los idiomas ibéricos. Los resultados revelan que, aunque algunos modelos destacan consistentemente en ciertas tareas, aún existen importantes brechas de rendimiento, particularmente para idiomas como el euskera. Estos hallazgos subrayan la necesidad de investigar más a fondo cómo equilibrar la compacidad del modelo con un rendimiento multilingüe robusto.}

\seplnkey{Large Language Models (LLMs), Benchmarking, Iberian Languages, Multilingual NLP, Compact Models, Edge Computing}

\seplnabstract{Large Language Models have significantly advanced natural language processing, achieving remarkable performance in tasks such as language generation, translation, and reasoning. However, their substantial computational requirements restrict deployment to high-end systems, limiting accessibility on consumer-grade devices. This challenge is especially pronounced for under-resourced languages like those spoken in the Iberian Peninsula, where relatively limited linguistic resources and benchmarks hinder effective evaluation. This work presents a comprehensive evaluation of compact state-of-the-art LLMs across several essential NLP tasks tailored for Iberian languages. The results reveal that while some models consistently excel in certain tasks, significant performance gaps remain, particularly for languages such as Basque. These findings highlight the need for further research on balancing model compactness with robust multilingual performance.}

\firstpageno{1}

\begin{document}


\setlength\titlebox{20cm} 

\label{firstpage} \maketitle

%

\section{Introduction}

Large Language Models (LLMs) have transformed natural language processing (NLP) by demonstrating state-of-the-art performance across a broad spectrum of tasks, ranging from language generation and translation to complex reasoning and comprehension. Notable examples are GPT-4 \cite{achiam2023gpt}, which excel in general reasoning and instruction-following capabilities, and Gemini 2 \cite{gemini2024}, which achieves remarkable translation accuracy. 

The continual increase in model scale and architectural complexity, often exceeding tens or even hundreds of billions of parameters, has produced significant advancements in linguistic proficiency, factual knowledge encoding, and context-aware interactions \cite{chiruzzo2024overview}. However, these advances come with substantial computational demands, requiring powerful GPUs and specialized server infrastructures. Consequently, access to these models is limited to users or organizations with significant computational resources, hindering broader adoption and deployment, especially in end-user devices or low-resource settings \cite{yuan2025llmpoweredbenchmarkfactoryreliable}.

In response to these resource constraints, recent NLP research has explored the development of compact or quantized LLMs that can efficiently run directly on consumer-grade hardware \cite{bucher2024fine}. The benefits of deploying models locally include improved response times, enhanced privacy due to reduced reliance on cloud infrastructure, and greater accessibility. In contrast to these advantages, compact models inherently pose trade-offs, balancing between their smaller requirements and the need to retain performance close to larger, computationally intensive counterparts. Evaluating the practical capabilities and limitations of these compact models in realistic usage scenarios thus becomes crucial for guiding both research directions and user expectations.

These challenges become particularly salient in the context of languages different from English, the main resource during training for most LLMs. Despite extensive research in multilingual NLP, there remains an evident gap concerning Iberian languages, namely Catalan, Spanish, Basque, Galician, and Portuguese \cite{chiruzzo2024overview}. While Spanish and Portuguese have historically received more attention, languages like Catalan, Basque, and Galician have comparatively fewer linguistic resources, datasets, and benchmarks. Recent literature \cite{chiruzzo2024overview} highlights the need for comprehensive evaluation frameworks to address specific linguistic nuances and computational challenges. Additionally, the rapid emergence of new compact models further complicates evaluations, as benchmarks must constantly adapt to maintain relevancy, underscoring the need for rigorous and continuously updated assessment strategies \cite{parashar2025inferencetimecomputationsllmreasoning}.

Recent work has begun developing resources and benchmarks to enhance the evaluation capabilities in multilingual NLP for Iberian languages. Notable efforts, such as Iberobench \cite{baucells2025iberobench}, offer promising starting points by introducing multiple tasks and systematically evaluating performance. However, despite these valuable initiatives, significant limitations remain. One crucial challenge is the rapid evolution and constant release of new LLMs, which leads to performance comparisons quickly becoming outdated or incomplete, as evaluations must continuously integrate newly available models. Additionally, subtler linguistic phenomena, such as irony, sarcasm, or complex inferential reasoning, remain understudied or assessed only superficially \cite{zhang2024sarcasmbench}. This gap leaves critical linguistic capabilities underexplored, limiting understanding of LLM performance and their practical usability in diverse and nuanced linguistic settings.

This paper presents a comprehensive evaluation of compact state-of-the-art LLMs across diverse NLP tasks designed for Iberian languages. To enable this evaluation, unified benchmarks are used, including Iberobench, covering multiple essential NLP tasks, and extending the irony detection dataset IroSvA \cite{Ortega2019IroSvA} to Catalan, Basque, Galician, and Portuguese. All models evaluated were publicly available as of March 2025. By benchmarking diverse compact LLMs across diverse tasks, this work identifies specific linguistic and task-related challenges, highlight performance gaps, and provide insights into the suitability of compact multilingual models for user deployment. 


\section{Related Work}
\label{sec:related}

\textbf{Multilingual Large Language Models} are a type of LLM trained on multiple languages, enabling them to understand and generate text in different languages. This capability makes them particularly useful for tasks such as translation. However, these models must balance their knowledge across multiple languages, often resulting in lower performance than monolingual models. A key challenge during training is ensuring that the data is balanced, as some languages are overrepresented. To address this issue, dominant languages, such as English, are downsampled, while underrepresented languages are upsampled. Another challenge for Multilingual LLMs is the tokenization, as different languages have distinct writing systems and word boundaries. The most used tokenizer for multilingual LLMs is Byte-Pair Encoding (BPE), an algorithm designed to compress texts to a predefined vocabulary size. 

As LLMs grow in size and complexity, their deployment becomes increasingly challenging. To address these issues, techniques such as \textbf{distillation} and \textbf{quantization} are widely used \cite{xu2024survey}. Distillation is a training method in which a ``student" model learns from the outputs of a larger ``teacher" model. These outputs can include the hard labels, the probability distribution, intermediate representations or the attention patterns of the teacher. Distilled models can be significantly smaller while maintaining most of the teacher model's performance.
Quantization is a technique mostly used during inference that reduces the numerical precision of model parameters to lower computational costs and decrease model size. Most LLMs use 32-bit floating-point precision to store their parameters; however, reducing the precision, for example to 4-bit integers, can shrink the model. While quantization improves efficiency, decreasing the precision makes the model less accurate, particularly in models used for complex reasoning tasks, calculations and other fine-grained language understanding applications. 

\textbf{Large Language Model evaluation} presents multiple challenges, as there is no definitive method to accurately measure their performance. Evaluation methods can be categorized into human evaluation, which is subjective and time-consuming, and automated evaluation, which is fast and scalable, but may not fully capture the nuances of meaning or inherent biases. Due to human evaluation limitations, most LLM performance assessments rely on automated evaluation with benchmarks such as AGIEval, ARC-Challenge, HumanEval, MMLU or WinoGrande. 
Even among widely used benchmarks, evaluation methods are not standardized across every model, as reflected in the \tablename~\ref{tab:benchmark_comparison}. Most reported results come from the original authors' publications, while others were obtained from external evaluations \cite{neuralmagic_deepseek_r1_distill_qwen_7b_quantized}. AGIEval is a benchmark designed to assess LLMs on problem-solving tasks, particularly real-world exams, and other high-stakes scenarios. ARC-Challenge and MMLU (Massive Multitask Language Understanding) evaluate models on common-sense reasoning and other capabilities using multiple choice questions. HumanEval focuses on measuring the coding capabilities of the models using programming problems. Finally, WinoGrande is a large-scale reasoning benchmark designed to evaluate models on pronoun resolution.

\begin{table}[ht!]
\footnotesize
\centering
\caption{Benchmark accuracy comparison}
\label{tab:benchmark_comparison}
\setlength{\tabcolsep}{2pt}
\resizebox{1.0\columnwidth}{!}{
\begin{tabular}{lccccc}
\toprule
\textbf{Model} & \textbf{\makecell[c]{AGI\\Eval}} & \textbf{\makecell[c]{ARC-\\Chall.}} & \textbf{\makecell[c]{Human\\Eval}} & \textbf{MMLU} & \textbf{Winogrande} \\
\midrule
DS-R1-Qwen-7B-it & - & 0.505 & - & 0.542 & 0.616 \\
Llama-3.1-8B-it & 0.478 & 0.797 & 0.726 & 0.667 & 0.605 \\
Ministral-3B-it & 0.421 & 0.642 & 0.342 & 0.609 & 0.727 \\
Ministral-8B-it & 0.483 & 0.719 & 0.348 & 0.650 & 0.753 \\
Mistral-7B-it & 0.425 & 0.642 & 0.268 & 0.625 & 0.742 \\
Qwen2.5-7B-it & - & 0.637 & 0.579 & 0.742 & 0.759 \\
Gemma 2-9B-it & 0.528 & 0.684 & 0.402 & 0.752 & 0.806 \\
Salamandra-2B-it & - & 0.354 & - & - & - \\
Salamandra-7B & - & 0.528 & - & - & - \\
Salamandra-7B-it & - & 0.545 & - & - & - \\
\midrule
Mistral Small 3-24B-it & - & - & 0.848 & 0.810 & - \\
\bottomrule
\end{tabular}
}
\end{table}

\section{Selected LLMs}
\label{sec:llms}

This section presents state-of-the-art LLMs with diverse designs and technical approaches, including multilingual and distilled variants. When available, instruction-tuned models are prioritized for evaluation.

\textbf{Mistral‑7B} \cite{jiang2023mistral7b}. This compact 7.3-billion-parameter decoder-only Transformer model is designed for efficiency and strong multilingual performance. It consists of 32 Transformer layers, each with a hidden dimension of 4096, and utilizes 32 attention heads, with 8 employing Grouped-Query Attention (GQA) for reduced memory usage during decoding. The architecture integrates rotary positional embeddings (RoPE) to efficiently encode token positions, RMS Norm for improved training and inference speed, and SwiGLU activations in the feed-forward network with a hidden dimension of 14336. Additionally, Mistral‑7B includes a sliding window attention (SWA) mechanism that limits attention computations within a fixed window, decreasing computational complexity for longer sequences.

\textbf{Ministral‑3B/8B} \cite{ministral8B}. Building upon Mistral‑7B, these models introduce several enhancements, notably extending the supported context length up to 128K tokens using an interleaved sliding-window attention with ragged segments (128k, 32k, 32k, 32k). They maintain RoPE and RMS Norm normalization to ensure efficient inference. Additionally, their feed-forward hidden dimension is increased to 12288, improving capacity for complex reasoning and instruction-following tasks. These refinements enable Ministral‑3B/8B to achieve improved multilingual and reasoning capabilities while maintaining efficiency.

\textbf{Salamandra 2B/7B} \cite{BSC_salamandra}. The Salamandra family represents a line of highly multilingual, decoder-only Transformer models developed at the Barcelona Supercomputing Center. The 7B variant has been trained on a corpus of curated data—spanning 35 European languages and code—delivering robust language generation and instruction-following capabilities. Additionally, the 2B model has a specialized Translation variant optimized for translation tasks, balancing a compact footprint with effective performance. These models incorporate advanced architectural features such as SwiGLU activations and rotary positional embeddings, making them well-suited for deployment in resource-constrained environments, including on-device inference.

\textbf{Llama-3.1-8B} \cite{dubey2024llama}. This is a member of Meta’s renowned Llama series, offering robust multilingual capabilities and versatile performance on a wide range of natural language processing tasks. With 8 billion parameters, this model strikes an effective balance between scale and efficiency, making it suitable for both research and practical applications such as dialogue, summarization, and general-purpose text generation. Architectural improvements over its predecessors, including refined training procedures and optimized parameter usage, ensure that Llama-3.1-8B delivers competitive performance while remaining lightweight enough for consumer-grade hardware.

\textbf{Qwen2.5-7B} \cite{yang2024qwen2}. It is a compact, 7-billion parameter language model that forms part of the Qwen series, known for its efficient handling of multilingual inputs and complex reasoning tasks. Designed with a focus on performance and low latency, Qwen2.5-7B incorporates innovative architectural optimizations that enable it to deliver high-quality text generation and inference on end-user devices. Its efficiency in balancing model size with output quality makes it particularly competitive in scenarios requiring swift, reliable responses in multiple languages, positioning it well in the emerging market of compact LLMs.

\textbf{DeepSeek-R1-Distill-Qwen-7B} \cite{guo2025deepseek}. This is a distilled model derived from the larger DeepSeek-R1 series, which was originally designed to excel in logical inference and complex reasoning tasks. By leveraging a comprehensive distillation process, where synthetic data generated by the full-scale DeepSeek-R1 model is used for supervised fine-tuning. This compact 7B-parameter Qwen model manages to retain much of the original reasoning prowess while being significantly more efficient. Its architecture is carefully optimized to deliver rapid inference on end-user devices, making it an attractive option for applications that demand a balance of performance and computational efficiency, particularly in domains like mathematical and language understanding.


\textbf{Gemma 2-9B} \cite{team2024gemma}. This is a 9.24‑billion‑parameter, decoder‑only Transformer model developed by Google. It employs an architecture that alternates between local sliding window attention (with an 8.2k token context window) and global attention in all other layers, using GQA for efficient processing. Built using the same research foundation as the Gemini family, Gemma 2 9B is designed for efficient text-to-text generation across tasks such as question answering, summarization, and reasoning. The model is optimized for deployment in resource‑constrained environments, supporting advanced quantization to balance performance and memory efficiency. With open weights available for both pre‑trained and instruction‑tuned variants, Gemma 2 9B democratizes access to state‑of‑the‑art natural language processing, making it suitable for both research and production settings.

\textbf{Mistral Small 3-24B} \cite{mistralsmall3}. Mistral‑24B scales the innovations of its smaller counterpart by increasing the model capacity to 24 billion parameters while preserving a latency-optimized, dense decoder‑only Transformer design. In effect, while retaining the efficiency and low‑latency characteristics of its predecessors, Mistral‑24B offers a noticeable uplift in performance, making it better suited for applications demanding both precision and speed without incurring additional computational overhead. This model is 4-bit quantized to fit in the available GPU.


\section{Iberian Language Tasks and Datasets}
\label{sec:benchmarks}

This section presents benchmarks designed to evaluate LLMs in Iberian languages. It first describes IberoBench, a multilingual benchmark covering several Iberian NLP tasks. Then, it introduces a dataset focused on irony and sarcasm detection.

\subsection{Iberobench benchmark}

For the evaluation of the performance of the Iberian languages, the Iberobench benchmark \cite{baucells2025iberobench} is chosen. This benchmark contains diverse datasets in five different languages: Catalan (ca), Spanish (es), Basque (eu), Galician (gl), and Portuguese (pt). Together, these datasets enable 62 different tasks grouped into 10 categories.

From the categories available, those with tasks in more languages are selected for benchmarking. Next, the categories used for the evaluation and their tasks are described.

\textbf{Reading Comprehension}: Belebele – A parallel reading comprehension task consisting of short passages with multiple-choice questions about text comprehension. Languages: ca, es, eu, gl, pt.

\textbf{Translation/Adaptation}: Flores – Translation tasks involving translations between Iberian languages and other European languages, based on the FLORES-200 dataset \cite{costa2022no}. Languages: ca, es, eu, gl, pt.

\textbf{Math}: MGSM\_direct - Mathematical reasoning tasks that consist of multilingual grade-school math word problems from the MGSM dataset \cite{shi2022language}. Languages: ca, es, eu, gl.

\textbf{Question Answering (QA)}: OpenbookQA – A dataset of questions that require multi-step reasoning and an understanding of scientific knowledge, translated professionally into Catalan, Spanish, and Galician.

\textbf{Natural Language Inference (NLI)}: WNLI, XNLI – Tasks involving natural language inference, where the models must determine logical relationships between pairs of sentences. Languages: ca, es, eu.

\textbf{Paraphrasing}: PAWS – Paraphrase identification tasks where models must identify if two sentences have the same meaning. Languages: ca, es, gl.

\subsection{Irony/Sarcasm Benchmark} 

To benchmark the capabilities of models in irony and sarcasm detection across Iberian languages, the Spanish subset of the IroSvA dataset \cite{Ortega2019IroSvA} was adapted into Catalan, Basque, Galician, and Portuguese. 

The IroSvA dataset comprises tweets from Spain and Mexico, as well as news comments from Cuba, annotated for irony based on specific contextual topics rather than isolated text. For this adaptation, only tweets from the Spain subset were selected to ensure regional linguistic coherence. The translation into Catalan, Basque, Galician, and Portuguese was carried out using Gemini 2 Flash \cite{gemini2024}, preserving the original contexts and annotation labels. The resulting multilingual dataset allows for evaluating performance in detecting irony and sarcasm across Iberian languages. 

Literal translation may introduce model specific biases, but the homogeneous Iberian context limits divergences in irony expression, so we focus on performance comparison. A comprehensive bias analysis would require replicating sociodemographic polarization methods beyond the scope of this work \cite{paper-multipico}.

\section{Results}
\label{sec:results}

This section presents the experimental outcomes of evaluating several compact LLMs on a diverse set of Iberian language tasks. Relevant scripts for data manipulation and evaluation are available at \cite{github}.

\subsection{Metrics and Methodology} 

Iberobench is included in the EleutherAI Language Model Evaluation Harness framework \cite{eval-harness}. Therefore, this framework is used in the evaluation. All tests are performed in a zero-shot fashion.

Accuracy was used as the evaluation metric for Belebele, OpenBookQA, PAWS, WNLI, and XNLI. These tasks involve multiple-choice or classification settings, such as selecting answers in reading comprehension and QA, identifying paraphrases, or determining logical relations in NLI tasks.

The metrics used for the machine translation benchmark Flores were the bilingual evaluation understudy (BLEU), Character-level F-score (ChrF), and the translation edit rate (TER). BLEU \cite{papineni2002bleu} is a metric used to evaluate the quality of a translation (higher is better), considering the correspondence between machine and human output. ChrF \cite{popovic2015chrf} is a metric that calculates the similarity between the the machine translation and a reference translation using the F1-score statistic (harmonic mean of precision and recall) for character n-gram matches (higher is better). TER \cite{snover2006study} quantifies the edit operations that a hypothesis requires to match a translation (lower is better). For evaluating the MGSM math benchmark, the exact\_match metric was used. This metric returns the rate at which the model exactly matches the expected output. For irony and sarcasm detection on the IroSvA dataset and its translations, accuracy and F1-score were used, as the models were prompted to classify messages as either ironic or non-ironic, effectively framing the task as a binary classification problem.

All evaluations were conducted using a modestly equipped server featuring an AMD EPYC 7313P CPU, an NVIDIA A10 GPU with 24 GB VRAM, and 132 GB of RAM. It is important to emphasize that this hardware setup is comparable to advanced consumer-grade computers. Thus, the reported results reflect realistic performance expectations for users without access to specialized or high-end computing resources.

\subsection{Evaluation Results}

\subsubsection{Reading Comprehension}

The results for machine reading comprehension performance on the Belebele dataset are shown in \tablename~\ref{tab:reading_comprehension}. The model Gemma 2-9B obtained the best results across all languages, achieving a mean accuracy of 0.87. This indicates its robust capability in understanding multilingual texts in Iberian languages. Closely following was the quantized model Mistral Small 3-24B, with a mean accuracy of 0.862, showing that quantized models with sufficient parameters can achieve competitive multilingual comprehension. Additionally, the Qwen2.5-7B and Ministral-8B models achieved reasonably good accuracies in Catalan, Galician, Portuguese, and Spanish; however, their performance dropped significantly for Basque, suggesting difficulties in handling languages that are more linguistically distant from Spanish or Portuguese. In contrast, the smaller models, such as Ministral-3B and SalamandraTA-2B, performed poorly across the board, highlighting that model size and complexity play essential roles in effectively handling multilingual reading comprehension tasks.

\begin{table}[ht!]
\footnotesize
\centering
\caption{Reading Comprehension Results}
\label{tab:reading_comprehension}
\setlength{\tabcolsep}{3pt}
\resizebox{1.0\columnwidth}{!}{
\begin{tabular}{lccccc}
\toprule
\multicolumn{6}{c}{\textbf{Dataset: Belebele}} \\
\multicolumn{6}{c}{\textbf{Metric: accuracy ($\uparrow$)}} \\
\midrule
\textbf{Model} & \textbf{ca} & \textbf{es} & \textbf{eu} & \textbf{gl} & \textbf{pt} \\
\midrule
DS-R1-Qwen-7B-it & 0.516 & 0.610 & 0.293 & 0.538 & 0.647 \\
Llama-3.1-8B-it & 0.638 & 0.690 & 0.432 & 0.646 & 0.720 \\
Ministral-3B-it & 0.264 & 0.274 & 0.287 & 0.266 & 0.268 \\
Ministral-8B-it & 0.756 & 0.804 & 0.550 & 0.770 & 0.803 \\
Mistral-7B-it & 0.667 & 0.637 & 0.330 & 0.592 & 0.673 \\
Qwen2.5-7B-it & 0.799 & 0.833 & 0.462 & 0.800 & 0.853 \\
\textbf{Gemma 2-9B-it} & \textbf{0.894} & \textbf{0.892} & \textbf{0.790} & \textbf{0.878} & \textbf{0.903} \\
Salamandra-2B-it &  0.298 & 0.241 & 0.287 & 0.227 & 0.261 \\
Salamandra-7B & 0.379 & 0.346 & 0.336 & 0.334 & 0.360 \\
Salamandra-7B-it & 0.674 & 0.652 & 0.573 & 0.637 & 0.639 \\
SalamandraTA-2B & 0.218 & 0.217 & 0.214 & 0.220 & 0.224 \\
\midrule
Mistral Small 3-24B-it & 0.880 & 0.891 & 0.779 & 0.874 & 0.894 \\
\bottomrule
\end{tabular}
}
\end{table}

\subsubsection{Translation and Adaptation}

\tablename~\ref{tab:translation} shows the results achieved in this task for the three selected metrics. The Mistral Small 3-24B excelled when evaluating it using BLEU  and ChrF, while Gemma 2-9B obtained better results when evaluating using TER. Both models obtained similar results, indicating strong translation quality, demonstrating their superior ability to generate fluent and accurate multilingual translations across Iberian languages. For this task, the third-best performing model was Llama-3.1-8B, which also obtained competitive results, although consistently lower than both Mistral Small 3-24B and Gemma. Together, this models significantly outperformed the others, highlighting a clear performance gap. The weakest-performing models, such as Ministral-3B and SalamandraTA-2B, struggled significantly, suggesting that smaller models may lack the required capacity or multilingual proficiency necessary for effective translation and adaptation tasks.

\begin{table}[ht!]
\centering
\scriptsize
\caption{Translation/Adaptation Results}
\label{tab:translation}
\resizebox{1.0\columnwidth}{!}{
\setlength{\tabcolsep}{2pt}
\begin{tabular}{lccccc}
\toprule
\multicolumn{6}{c}{\textbf{Dataset: Flores}} \\
\multicolumn{6}{c}{\textbf{Metric: BLEU ($\uparrow$)}} \\
\midrule
\textbf{Model} & \textbf{ca} & \textbf{es} & \textbf{eu} & \textbf{gl} & \textbf{pt} \\
\midrule
DS-R1-Qwen-7B-it & 0.088 & 0.099 & 0.009 & 0.079 & 0.133 \\
Llama-3.1-8B-it & 0.260 & 0.211 & 0.090 & 0.232 & 0.277 \\
Ministral-3B-it & 0.001 & 0.001 & $\approx$ 0 & $\approx$ 0 & 0.001 \\
Ministral-8B-it & 0.231 & 0.185 & 0.073 & 0.220 & 0.273 \\
Mistral-7B-it & 0.241 & 0.190 & 0.024 & 0.172 & 0.248 \\
Qwen2.5-7B-it & 0.195 & 0.193 & 0.016 & 0.190 & 0.253 \\
Gemma 2-9B-it & 0.290 & 0.233 & 0.135 & 0.262 & 0.314 \\
Salamandra-2B-it & 0.159 & 0.130 & 0.072 & 0.146 & 0.156 \\
Salamandra-7B & 0.104 & 0.076 & 0.041 & 0.106 & 0.118 \\
Salamandra-7B-it & 0.175 & 0.165 & 0.081 & 0.182 & 0.191 \\
SalamandraTA-2B & $\approx$ 0 & $\approx$ 0 & $\approx$ 0 & $\approx$ 0 & $\approx$ 0 \\
\midrule
\textbf{Mistral Small 3-24B-it} & \textbf{0.301} & \textbf{0.237} & \textbf{0.145} & \textbf{0.276} & \textbf{0.322} \\

\toprule
\multicolumn{6}{c}{\textbf{Dataset: Flores}} \\
\multicolumn{6}{c}{\textbf{Metric: ChrF ($\uparrow$)}} \\
\midrule
\textbf{Model} & \textbf{ca} & \textbf{es} & \textbf{eu} & \textbf{gl} & \textbf{pt} \\
\midrule
DS-R1-Qwen-7B-it & 0.358 & 0.373 & 0.188 & 0.345 & 0.412 \\
Llama-3.1-8B-it & 0.543 & 0.510 & 0.385 & 0.521 & 0.557 \\
Ministral-3B-it & 0.071 & 0.080 & 0.061 & 0.068 & 0.074 \\
Ministral-8B-it & 0.506 & 0.470 & 0.351 & 0.506 & 0.546 \\
Mistral-7B-it & 0.518 & 0.484 & 0.233 & 0.467 & 0.524 \\
Qwen2.5-7B-it & 0.471 & 0.473 & 0.150 & 0.470 & 0.516 \\
Gemma 2-9B-it & 0.573 & 0.532 & 0.445 & 0.553 & 0.589 \\
Salamandra-2B-it & 0.454 & 0.424 & 0.393 & 0.448 & 0.445 \\
Salamandra-7B & 0.344 & 0.325 & 0.260 & 0.359 & 0.369 \\
Salamandra-7B-it & 0.490 & 0.479 & 0.430 & 0.501 & 0.513 \\
SalamandraTA-2B & 0.065 & 0.077 & 0.066 & 0.057 & 0.065 \\
\midrule
\textbf{Mistral Small 3-24B-it} & \textbf{0.583} & \textbf{0.537} & \textbf{0.473} & \textbf{0.566} & \textbf{0.596} \\

\toprule
\multicolumn{6}{c}{\textbf{Dataset: Flores}} \\
\multicolumn{6}{c}{\textbf{Metric: TER ($\downarrow$)}} \\
\midrule
\textbf{Model} & \textbf{ca} & \textbf{es} & \textbf{eu} & \textbf{gl} & \textbf{pt} \\
\midrule
DS-R1-Qwen-7B-it & 108.54 & 101.77 & 202.56 & 116.69 & 96.93 \\
Llama-3.1-8B-it & 67.43 & 71.08 & 112.66 & 71.43 & 64.36 \\
Ministral-3B-it & 485.44 & 448.44 & 516.05 & 507.63 & 483.41 \\
Ministral-8B-it & 72.24 & 75.26 & 114.62 & 72.71 & 64.39 \\
Mistral-7B-it & 71.30 & 75.73 & 143.43 & 79.53 & 69.10 \\
Qwen2.5-7B-it & 204.16 & 200.05 & 1181.98 & 201.60 & 189.45 \\
\textbf{Gemma 2-9B-it} & 60.05 & \textbf{65.98} & \textbf{84.57} & 62.59 & \textbf{57.46} \\
Salamandra-2B-it & 80.29 & 82.16 & 95.61 & 79.37 & 79.67 \\
Salamandra-7B & 81.34 & 84.36 & 96.77 & 80.02 & 78.85 \\
Salamandra-7B-it & 113.59 & 82.62 & 150.40 & 89.04 & 99.75 \\
SalamandraTA-2B & 346.78 & 346.80 & 330.16 & 309.62 & 402.23 \\
\midrule
Mistral Small 3-24B-it & \textbf{59.31} & 66.38 & 89.64 & \textbf{61.72} & 57.51 \\
\bottomrule
\end{tabular}
}
\end{table}

\subsubsection{Math}

\tablename~\ref{tab:math} shows the results when evaluating the math performance on the MGSM\_direct dataset. The model that obtained the best overall results was Qwen2.5-7B, achieving a mean exact match of 1.57. It notably excelled in Catalan, Spanish, and Galician, achieving particularly high performance in Spanish (0.390 exact match), while maintaining moderate accuracy in Catalan (0.096) and Galician (0.100). However, for Basque, the quantized model Mistral Small 3-24B significantly outperformed all others, achieving the highest exact match score (0.134), despite lower results in the remaining languages. This highlights the unique challenge Basque poses to general-purpose models, possibly due to its linguistic distinctiveness compared to the other Iberian languages. Again, smaller model variants performed poorly across all languages, indicating their insufficient capacity for complex mathematical reasoning tasks.

\begin{table}[ht!]
\centering
\scriptsize
\caption{Math Results}
\label{tab:math}
\setlength{\tabcolsep}{3pt}
\begin{tabular}{lcccc}
\toprule
\multicolumn{5}{c}{\textbf{Dataset: MGSM\_direct}}\\ 
\multicolumn{5}{c}{\textbf{Metric: exact\_match ($\uparrow$)}} \\
\midrule
\textbf{Model} & \textbf{ca} & \textbf{es} & \textbf{eu} & \textbf{gl} \\
\midrule
DS-R1-Qwen-7B-it & 0.032 & 0.110 & 0.040 & 0.012 \\
Llama-3.1-8B-it & 0.004 & 0.128 & 0.034 & 0.000 \\
Ministral-3B-it & 0.008 & 0.010 & 0.006 & 0.016 \\
Ministral-8B-it & 0.012 & 0.290 & 0.048 & 0.008 \\
Mistral-7B-it & 0.000 & 0.136 & 0.012 & 0.000 \\
\textbf{Qwen2.5-7B-it} & \textbf{0.096} & \textbf{0.390} & 0.042 & \textbf{0.100} \\
Gemma 2-9B-it & 0.032 & 0.368 & 0.000 & 0.000 \\
Salamandra-2B-it & 0.004 & 0.018 & 0.006 & 0.000 \\
Salamandra-7B & 0.008 & 0.012 & 0.014 & 0.016 \\
Salamandra-7B-it & 0.004 & 0.020 & 0.020 & 0.000 \\
\midrule
Mistral Small 3-24B-it & 0.036 & 0.310 & \textbf{0.134} & 0.040 \\

\bottomrule
\end{tabular}
\end{table}

\subsubsection{Question Answering}

The dataset OpenbookQA was used to evaluate the question-answering task, with results detailed in \tablename~\ref{tab:qa}. The instruction-tuned Salamandra-7B model obtained the highest accuracy consistently across all languages, achieving a mean accuracy of 0.38. This highlights the effectiveness of instruction-tuning for knowledge-intensive QA tasks. The second-best model was Gemma 2-9B, with a lower mean accuracy of 0.338, yet still demonstrating strong multilingual capabilities. In general, Spanish was the language in which models performed best, followed by Catalan and Galician. The smaller models again showed limited performance, suggesting that their size and lack of tuning for complex QA tasks negatively affected the results.

\begin{table}[ht!]
\centering
\footnotesize
\caption{QA Results}
\label{tab:qa}
\begin{tabular}{lcccc}
\toprule
\multicolumn{4}{c}{\textbf{Dataset: OpenbookQA}}\\
\multicolumn{4}{c}{\textbf{Metric: accuracy ($\uparrow$)}} \\
\midrule
\textbf{Model} & \textbf{ca} & \textbf{es} & \textbf{gl} \\
\midrule
DS-R1-Qwen-7B-it & 0.204 & 0.238 & 0.246 \\
Llama-3.1-8B-it & 0.282 & 0.352 & 0.294 \\
Ministral-3B-it & 0.216 & 0.198 & 0.202 \\
Ministral-8B-it & 0.280 & 0.360 & 0.308 \\
Mistral-7B-it & 0.284 & 0.330 & 0.272 \\
Qwen2.5-7B-it & 0.272 & 0.334 & 0.272 \\
Gemma 2-9B-it & 0.350 & 0.356 & 0.308 \\
Salamandra-2B-it & 0.290 & 0.310 & 0.252 \\
Salamandra-7B & 0.310 & 0.348 & 0.300 \\
\textbf{Salamandra-7B-it} & \textbf{0.364} & \textbf{0.392} & \textbf{0.328} \\
\midrule
Mistral Small 3-24B-it & 0.346 & 0.356 & 0.306 \\
\bottomrule
\end{tabular}
\end{table}

\subsubsection{Natural Language Inference}

The performance of models on NLI tasks was evaluated using the WNLI and XNLI datasets, as shown in \tablename~\ref{tab:nli_wnli}. The best-performing model on the WNLI dataset was Gemma 2-9B, with notably high accuracies in Catalan (0.775) and Spanish (0.789), but weaker results in Basque (0.465). In contrast, for the XNLI dataset, the Salamandra-7B model achieved the highest mean accuracy, demonstrating more balanced performance across the three languages evaluated. This variation highlights the task and language-specific differences across models.

\begin{table}[ht!]
\centering
\footnotesize
\caption{NLI Results}
\label{tab:nli_wnli}
\begin{tabular}{lccc}
\toprule
\multicolumn{4}{c}{\textbf{Dataset: WNLI}} \\
\multicolumn{4}{c}{\textbf{Metric: accuracy ($\uparrow$)}} \\
\midrule
\textbf{Model} & \textbf{ca} & \textbf{es} & \textbf{eu} \\
\midrule
DS-R1-Qwen-7B-it & 0.535 & 0.563 & 0.493 \\
Llama-3.1-8B-it & 0.437 & 0.493 & 0.563 \\
Ministral-3B-it & 0.535 & 0.479 & 0.563 \\
Ministral-8B-it & 0.437 & 0.634 & 0.437 \\
Mistral-7B-it & 0.507 & 0.592 & 0.592 \\
Qwen2.5-7B-it & 0.563 & 0.704 & 0.521 \\
\textbf{Gemma 2-9B-it} & \textbf{0.775} & \textbf{0.789} & 0.465 \\
Salamandra-2B-it & 0.563 & 0.451 & 0.437 \\
Salamandra-7B & 0.549 & 0.423 & 0.606 \\
Salamandra-7B-it & 0.592 & 0.437 & 0.521 \\
\midrule
Mistral Small 3-24B-it & 0.578 & 0.718 & \textbf{0.620} \\
\midrule

\multicolumn{4}{c}{\textbf{Dataset: XNLI}} \\
\multicolumn{4}{c}{\textbf{Metric: accuracy ($\uparrow$)}} \\
\midrule
\textbf{Model} & \textbf{ca} & \textbf{es} & \textbf{eu} \\
\midrule
DS-R1-Qwen-7B-it & 0.403 & 0.442 & 0.334 \\
Llama-3.1-8B-it & 0.513 & 0.494 & 0.411 \\
Ministral-3B-it & 0.335 & 0.336 & 0.336 \\
Ministral-8B-it & 0.527 & 0.466 & 0.425 \\
Mistral-7B-it & 0.508 & 0.439 & 0.315 \\
Qwen2.5-7B-it & 0.498 & 0.482 & 0.400 \\
Gemma 2-9B-it & 0.513 & 0.488 & 0.449 \\
Salamandra-2B-it & 0.486 & 0.466 & 0.430 \\
\textbf{Salamandra-7B} & 0.537 & \textbf{0.507} & 0.456 \\
Salamandra-7B-it & 0.531 & 0.480 & \textbf{0.475} \\
\midrule
Mistral Small 3-24B-it & \textbf{0.540} & 0.497 & 0.455 \\
\bottomrule

\end{tabular}
\end{table}

\subsubsection{Paraphrasing}

The paraphrasing task, evaluated using the PAWS dataset, revealed mixed results among top-performing models, as detailed in \tablename~\ref{tab:paraphrasing}. The model with the highest accuracy in Spanish was Mistral-7B (0.689), whereas the model Mistral Small 3-24B achieved better performance in both Catalan (0.706) and Galician (0.678). Despite this language-specific variation, Mistral-7B achieved the highest overall mean accuracy (0.683), highlighting its balanced multilingual proficiency in detecting paraphrases. Other models, such as Gemma 2-9B and Qwen2.5-7B, also performed well, but with less consistency across languages. In contrast, the smaller Salamandra-2B and DS-R1-Qwen-7B models performed notably poorer, underscoring the importance of model capacity in handling nuanced paraphrase identification tasks.

\begin{table}[ht!]
\footnotesize
\centering
\caption{Paraphrasing Results}
\label{tab:paraphrasing}
\begin{tabular}{lccc}
\toprule
\multicolumn{4}{c}{\textbf{Dataset: PAWS}}\\
\multicolumn{4}{c}{\textbf{Metric: accuracy ($\uparrow$)}} \\
\midrule
\textbf{Model} & \textbf{ca} & \textbf{es} & \textbf{gl} \\
\midrule
DS-R1-Qwen-7B-it & 0.599 & 0.550 & 0.551 \\
Llama-3.1-8B-it & 0.655 & 0.652 & 0.645 \\
Ministral-3B-it & 0.571 & 0.583 & 0.584 \\
Ministral-8B-it & 0.658 & 0.643 & 0.661 \\
\textbf{Mistral-7B-it} & 0.685 & \textbf{0.689} & 0.676 \\
Qwen2.5-7B-it & 0.664 & 0.651 & 0.605 \\
Gemma 2-9B-it & 0.679 & 0.669 & 0.629 \\
Salamandra-2B-it & 0.576 & 0.560 & 0.535 \\
Salamandra-7B & 0.606 & 0.609 & 0.576 \\
Salamandra-7B-it & 0.675 & 0.606 & 0.608 \\
\midrule
Mistral Small 3-24B-it & \textbf{0.706} & 0.663 & \textbf{0.678} \\
\bottomrule
\end{tabular}
\end{table}

\subsubsection{Irony and Sarcasm Detection}

The irony and sarcasm detection task revealed clear differences in model performance across languages and architectures, as shown in \tablename~\ref{tab:irony}. Some models consistently predicted the majority class (“No”), leading to misleadingly high accuracy due to the class imbalance. As a result, F1-score is reported instead, which better reflects performance across both classes. The best overall performance was achieved by Mistral Small 3-24B, which obtained the highest or near-highest F1 scores in all five languages, including 0.52 in Catalan and Spanish, and 0.53 in Galician. For Basque and Portuguese, its F1 scores remained high (0.48 and 0.51, respectively), indicating robust cross-lingual generalization in irony detection. Qwen2.5-7B and Gemma 2-9B also delivered strong results, with the latter showing highly consistent F1 scores (close to 0.49) in the five languages. Basque emerged as the most challenging language, with lower scores in most models, while Galician and Portuguese generally showed higher performance. Given that irony and sarcasm are deeply influenced by language and culture, the translated benchmark may not fully convey the intended meaning of each message. However, the consistent results across languages suggest the translations effectively preserved these nuances. These results emphasize the difficulty of irony detection and the value of larger, better-tuned models for achieving robust multilingual performance.

\begin{table}[ht!]
\centering
\footnotesize
\caption{Irony and Sarcasm Results}
\label{tab:irony}
\resizebox{1.0\columnwidth}{!}{
\setlength{\tabcolsep}{3pt}
\begin{tabular}{lccccc}
\toprule
\multicolumn{6}{c}{\textbf{Dataset: IroSvA}} \\
\multicolumn{6}{c}{\textbf{Metric: accuracy ($\uparrow$)}} \\
\midrule
\textbf{Model} & \textbf{ca} & \textbf{es} & \textbf{eu} & \textbf{gl} & \textbf{pt} \\
\midrule
DS-R1-Qwen-7B-it & \textbf{0.616} & 0.463 & 0.121 & 0.362 & 0.182 \\
Llama-3.1-8B-it & 0.471 & 0.464 & 0.258 & 0.346 & 0.440 \\
Ministral-3B-it & 0.424 & 0.443 & 0.059 & 0.499 & 0.454 \\
Ministral-8B-it & 0.420 & \textbf{0.658} & 0.340 & \textbf{0.657} & 0.547 \\
Mistral-7B-it & 0.411 & 0.598 & 0.076 & 0.339 & 0.443 \\
Qwen2.5-7B-it & 0.411 & 0.544 & \textbf{0.621} & 0.546 & \textbf{0.589} \\
Gemma 2-9B-it & 0.45 & 0.356 & 0.544 & 0.372 & 0.510 \\
Salamandra-2B-it & 0.579 & 0.526 & 0.481 & 0.468 & 0.485 \\
Salamandra-7B & 0.432 & 0.598 & 0.384 & 0.449 & 0.583 \\
Salamandra-7B-it & 0.334 & 0.365 & 0.379 & 0.319 & 0.354 \\
\midrule
Mistral Small 3-24B-it & 0.458 & 0.502 & 0.576 & 0.552 & 0.579 \\
\toprule
\multicolumn{6}{c}{\textbf{Dataset: IroSvA}} \\
\multicolumn{6}{c}{\textbf{Metric: F1-score ($\uparrow$)}} \\
\midrule
\textbf{Model} & \textbf{ca} & \textbf{es} & \textbf{eu} & \textbf{gl} & \textbf{pt} \\
\midrule
DS-R1-Qwen-7B-it & 0.150 & 0.338 & 0.031 & 0.457 & 0.244 \\
Llama-3.1-8B-it & 0.383 & 0.410 & 0.204 & 0.439 & 0.340 \\
Ministral-3B-it & 0.007 & 0.000 & 0.002 & 0.000 & 0.000 \\
Ministral-8B-it & 0.522 & 0.290 & \textbf{0.499} & 0.023 & \textbf{0.541} \\
Mistral-7B-it & 0.516 & 0.487 & 0.009 & 0.501 & 0.520 \\
Qwen2.5-7B-it & 0.490 & 0.500 & 0.117 & 0.430 & 0.460 \\
Gemma 2-9B-it & 0.467 & 0.466 & 0.490 & 0.485 & 0.493 \\
Salamandra-2B-it & 0.233 & 0.303 & 0.162 & 0.403 & 0.360 \\
Salamandra-7B & 0.380 & 0.242 & 0.477 & 0.472 & 0.263 \\
Salamandra-7B-it & 0.483 & 0.450 & 0.457 & 0.459 & 0.451 \\
\midrule
\textbf{Mistral Small 3-24B-it} & \textbf{0.523} & \textbf{0.517} & 0.477 & \textbf{0.529} & 0.509 \\
\bottomrule
\end{tabular}
}
\end{table}

\section{Discussion}
\label{sec:discussion}

This section examines cross-task performance patterns, efficiency trade-offs, and language-specific challenges.

\subsection{Distillation Impact}

DeepSeek-R1-Qwen-7B consistently underperforms compared to Qwen2.5-7B across all tasks, except in Basque translation task measured by TER. This underperformance may be explained by the fact that the student model, based in Qwen2.5-Math-7B, is not instruction-tuned and this likely limited its ability to inherit knowledge during distillation. Besides, the base model is focused on reasoning on English and Chinese.

The decreased performance in Basque can also be attributed to the distillation process removing specialized performance in that language domain while improving broader generalization. Additionally, DeepSeek-R1-Qwen-7B produces shorter, more literal translations closely matching reference texts, resulting in lower TER. In contrast, Qwen2.5-7B generates more fluent, semantically rich translations, achieving higher BLEU and ChrF scores but also higher TER due to structural variations from the reference.

\subsection{Instruction Tuning Impact}

\begin{figure}[htpb!]
    \centering
    \includegraphics[width=\linewidth, trim=0 10 0 10, clip]{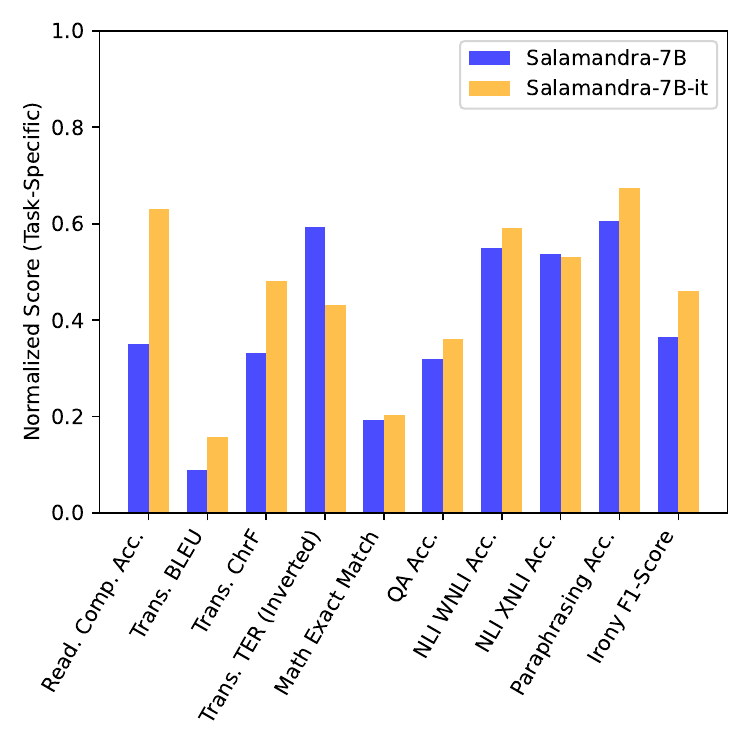}
    \caption{Salamandra-7B comparison}
    \label{fig:salamandra}
\end{figure}

Salamandra-7B with and without instruction tuning were evaluated to assess how instruction tuning influences model performance. \figurename~\ref{fig:salamandra} shows the normalized score for each evaluated task. Note that TER is inverted and normalization ranges were different depending on the task.

The instruction-tuned variant consistently outperformed the base Salamandra-7B model in reading comprehension (mean accuracy increased from approximately 0.35 to 0.63), translation (average BLEU increased notably from 0.089 to 0.159), and question answering (mean accuracy improved from 0.319 to 0.361). These gains were particularly pronounced in Spanish and Portuguese, where the model benefits from richer training resources, while improvements in Basque and Galician were more limited, likely due to their lower representation in training corpora.

In contrast, for tasks such as paraphrasing and natural language inference, instruction tuning provided mixed results, sometimes yielding marginal improvements or even slight performance declines. These findings suggest that instruction tuning benefits knowledge-intensive and comprehension tasks, but its impact may vary depending on task complexity and linguistic demands.

\subsection{Execution Time Analysis}

\tablename~\ref{tab:exec_times} reports the execution time for the reading comprehension tests (Belebele datasets). Smaller models, such as Mistral-3B and Salamandra-2B, completed the tests significantly faster. In contrast, larger models required more time to complete their tests. Particularly, Mistral Small 3-24B, even after quantization, took up to three times longer than other models such as Gemma 2-9B or Llama3.1-8B. Language-wise,  Basque tests took $\approx$7-15\% longer than those in other languages for most models, except when using Salamandra models, which were specifically trained on Iberian Languages.

\begin{table}[ht!]
\centering
\footnotesize
\caption{Reading comp. execution times}
\label{tab:exec_times}
\setlength{\tabcolsep}{4pt}
\begin{tabular}{lccccc}
\toprule
\multicolumn{6}{c}{\textbf{Dataset: Belebele}} \\
\multicolumn{6}{c}{\textbf{Metric: time (seconds)}} \\
\midrule
\textbf{Model} & \textbf{ca} & \textbf{es} & \textbf{eu} & \textbf{gl} & \textbf{pt} \\
\midrule
DS-R1-Qwen-7B-it & 101 & 96 & 109 & 104 & 96 \\
Llama-3.1-8B-it & 102 & 96 & 111 & 99 & 95 \\
Ministral-3B-it & 56 & 56 & 65 & 57 & 56 \\
Ministral-8B-it & 94 & 92 & 100 & 94 & 93 \\
Mistral-7B-it & 102 & 104 & 117 & 105 & 102 \\
Qwen2.5-7B-it & 101 & 96 & 109 & 99 & 95 \\
Gemma 2-9B-it & 128 & 115 & 141 & 120 & 115 \\
Salamandra-2B-it & 42 & 42 & 41 & 42 & 42 \\
Salamandra-7B & 86 & 84 & 81 & 82 & 82 \\
Salamandra-7B-it & 85 & 84 & 81 & 83 & 82 \\
SalamandraTA-2B & 45 & 42 & 41 & 41 & 41 \\
\midrule
Mistral Small 3-24B-it & 348 & 342 & 366 & 348 & 348 \\
\bottomrule
\end{tabular}
\end{table}

\subsection{Basque Analysis} 

\begin{figure}[htpb!]
    \centering
    \includegraphics[width=\linewidth, trim=0 10 0 10, clip]{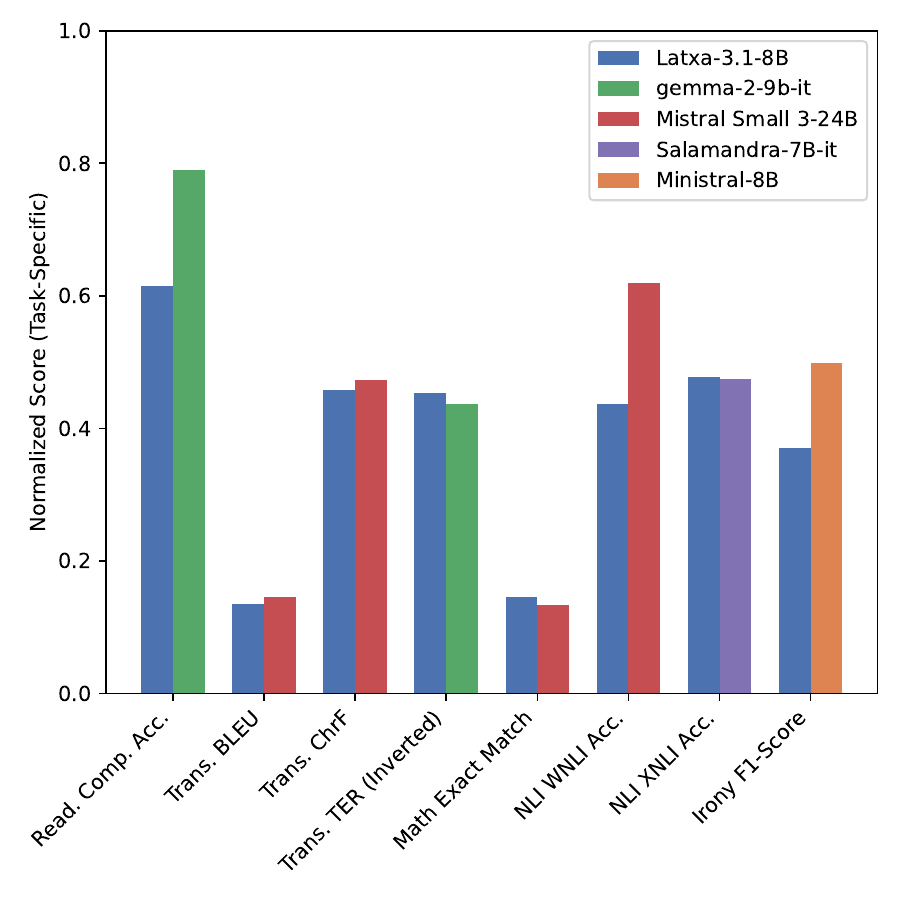}
    \caption{Latxa comparison}
    \label{fig:latxa}
\end{figure}

The Basque language poses a unique challenge for NLP, as it is a language isolate with no known relatives. LLMs often leverage similarities among related languages, such as the Romance family, to generalize knowledge. In contrast, LLMs must learn Basque almost entirely from scratch, making the training process more demanding. \cite{etxaniz2024latxaopenlanguagemodel} proposed a dedicated Basque corpus for training and evaluating language models, which was used to develop the model Latxa-Llama-3.1-8B-Instruct, a Llama-3.1-8B adaptation for Basque. \figurename~\ref{fig:latxa} shows the comparison for Basque between Latxa and the strongest competing model in each task. Latxa underperforms in Reading Comprehension, Translation (evaluated using BLEU and ChrF), NLI on the WNLI benchmark and Irony Detection. However, it performs slightly better in Translation when evaluated with TER, as well as Math and NLI on the XNLI benchmark.

Despite trained specifically for Basque, Latxa does not overperforms compared to similar size models, such as Gemma 2-9B or Ministral-8B. If all benchmark scores are normalized and averaged, Latxa is placed third after Mistral-3-24B and Gemma-2-9B. This may be partially attributed to the performance of its base model, which did not achieve top results across benchmarks, either in average multilingual performance or for Basque only. Still, Latxa-3.1-8B improves upon Llama-3.1-8B, proving that specialized models iterated on monolingual data can achieve better performance.

\section{Conclusions and Future Work}
\label{sec:conclusion}

This study evaluated compact LLMs on a range of Iberian language tasks using unified benchmarks such as Iberobench and extended datasets such as IroSvA. Tasks included reading comprehension, translation, mathematical reasoning, question answering, natural language inference, and paraphrasing across Catalan, Spanish, Basque, Galician, and Portuguese. The results indicate that current compact LLMs can achieve competitive performance in multilingual settings. For instance, Gemma 2‑9B excelled in reading comprehension and translation, while instruction‑tuned variants performed well in question answering. However, challenges persist, particularly for under‑resourced languages like Basque, and quantized models sometimes exhibit trade-offs in accuracy compared to full‑precision versions.

Future research should extend evaluations to new LLMs and additional tasks capturing subtler linguistic phenomena, such as sentiment and pragmatic reasoning. Efforts should also focus on improving performance for under‑resourced languages through targeted pretraining and fine‑tuning, and on refining deployment strategies for resource‑constrained environments. Future work will also incorporate more statistically robust evaluation methods to ensure the reliability and significance of comparisons.

\section*{Acknowledgements} 

This work has been funded by the European Union - Next Generation EU, through the Investigo program with file number 09-PIN1-00054.1/2023 of the Consejería de Economía, Hacienda y Empleo de la Comunidad de Madrid.

\bibliographystyle{fullname}
\bibliography{EjemploARTsepln}

\appendix

\end{document}